# An Open-Source Scenario Architect for Autonomous Vehicles

Tim Stahl, Johannes Betz
Chair of Automotive Technology
Technical University of Munich
Garching, Germany
Email: tim.stahl@tum.de

*Abstract*—The development of software components for autonomous driving functions should always include an extensive and rigorous evaluation. Since real-world testing is expensive and safety-critical – especially when facing dynamic racing scenarios at the limit of handling – a favored approach is simulation-based testing. In this work, we propose an open-source graphical user interface, which allows the generation of a multi-vehicle scenario in a regular or even a race environment. The underlying method and implementation is elaborated in detail. Furthermore, we showcase the potential use-cases for the scenario-based validation of a safety assessment module, integrated into an autonomous driving software stack. Within this scope, we introduce three illustrative scenarios, each focusing on a different safety-critical aspect.

*Keywords*—*Autonomous vehicles; safety assessment; scenario-based testing; scenario architect.*

## I. Introduction

To ensure the safety of developed software components for autonomous driving functions, a comprehensive and rigorous evaluation is always carried out. The more complex a function under development is, the more effort and expense has to be devoted to testing. Recent studies [1], [2] claim that the approval of an autonomous vehicle requires several million test-kilometers to be driven.

Since real-world testing is expensive and safety-critical, a preferred approach is scenario-based testing. In our research [3], we deal with autonomous racing vehicles, where – due to the fact that such cars are driven at the limit of handling – this aspect is even more important. The appropriate selection of challenging scenarios can decrease the testing burden further. However, the proper design and implementation of such scenarios can be cumbersome. In this work, we propose a graphical user interface, which allows a user to generate a multi-vehicle scenario for a race environment. The velocity profile for each user-generated path is initialized in a race-realistic manner. Furthermore, we showcase the generation of fault-injected scenarios (e.g. intended collisions) by detailing three examples and outlining the practicability for continuous integration (CI). The tool will be publicly available via GitHub (github.com/TUMFTM/ScenarioArchitect).

For approval of state-of-the-art advanced driver assistance systems, norms like the ISO 26262 [4] are pursued during development and final field tests – requested by the ECE-homologation process – have to be met. With higher levels of automation, the complexity of systems increases, and having the human as a fall-back strategy is removed. Therefore, much more effort is expended in testing and approval. According to recent studies [1], [2], several million test-kilometers are required in order to confidently claim safer behavior when compared to a human counterpart.

Original equipment manufacturers (OEMs) and researchers are establishing datasets based on real driving data to be used for training and approval purposes. Waymo recently published a dataset with 1000 diverse scenarios, each running to around 20 seconds in length [5]. Aside from this, there are several other well-known and widely used datasets available, in the research domain. Among them is the Kitti dataset [6], primarily focusing on object detection, and the highD dataset [7], which provides a third person view, useful for motion planning and risk assessment research.

In all of the present datasets, critical scenes or situations resulting in a crash are rare or not present at all. However, the behavior of the vehicle under test (VUT) in such situations is crucial for approval purposes. Furthermore, to the best of the authors' knowledge, there is currently no dataset available which covers dynamic interactions or critical situations in a race environment.

In order to reduce the testing burden, some authors suggest scenario-based evaluation [8]. In this domain, scenarios are laid out by experts, based on real driving data or algorithmic selection and generation metrics [9], [10]. By focusing on challenging or relevant scenarios, the number of required scenarios for approval can be reduced further.

CommonRoad [11] is a collection of scenarios specifically for the purpose of benchmarking trajectory planners. The framework contains datasets based on real driving data as well as hand-crafted hazardous scenarios. However, the ego-trajectory is not provided in the samples. Therefore, dangerous situations are limited to dynamic collision risks, highly depended on the trajectory planner inserted into framework.

Especially in the domain of approval, artificially designed scenarios are the focus of most attention. Using this method, it is possible to generate scenarios according to customized requirements, which would be too expensive or safety-critical to test in real-life. Instead of implementing every scenario by hand, graphical user interfaces support the developer in establishing scenario datasets with acceptable effort. To the best of the authors' knowledge, there are currently only a few scenario-design tools available. MathWorks's commercial 'Automated Driving Toolbox' offers a 2D scenario designer [12] that supports basic road layouts with multiple vehicles and constant velocity-profile initializations. Vires' commercial 'VTD - Vires Virtual Test Drive' [13] provides several stages of design tools, including a 2D scenario editor mainly focusing on the path-layout than on the velocity-profile initialization and manipulation.

Scenario-based testing has become a crucial validation method. In this work, we present the following advantages compared to existing scenario-generation tools.

1) Open-source and free of charge, usable on common platforms (Windows, Linux, macOS).
2) Suitable for modeling various driving constellations, including race scenarios.
3) Plausible velocity profile initialization based on friction (constant velocity in existing work).
4) Intuitive graphical velocity profile manipulation along the course (respecting friction potential and including batch processing).
5) Live trajectory and status visualization for all vehicles, supporting an intuitive scenario design.
6) Framework and exemplary scenarios targeting automated scenario-based evaluation.

This paper is organized as follows. The basic functionality, as well as the physical notion describing the motion of the vehicle under test (VUT) and other traffic participants, are reviewed in Section II. Testing pipelines and evaluation results are summarized in Section III. Section IV provides a discussion and conclusion.

## II. Scenario Architect – Front- and Back-End

The proposed Scenario Architect offers a graphical user interface (GUI) that displays relevant information and allows for direct user inputs. Furthermore, the back-end processes all the data, calculates the path splines accompanied with velocity profiles and allows for data import and export. The GUI and key back-end components are elaborated on in subsequent subsections.

### A. GUI

The graphical interface consists of two windows (Fig. 1). The main window holds a 2D grid and key interface elements like buttons, radio-buttons and sliders. The second window holds all the temporal information for the current scene. Thereby, each object's velocity and acceleration course are plotted against time and distance.

The plot on the main window (Fig. 1a) displays one single time instance of the scenario in a bird's-eye view. The track is displayed in gray with bounds marked in black. Each vehicle is represented by a rectangle with the dimensions of the vehicle footprint and a unique color. The path a vehicle will travel for the duration of the scenario is visualized by a thin line of the corresponding color. Additionally, the trajectory of the VUT (orange) for the current time-stamp based on a configurable planning horizon, is highlighted by a solid red line.

The second window (Fig. 1b) visualizes temporal information regarding the scenario. Therefore, the velocity and acceleration course of every vehicle in the scene is plotted. The top plot displays lateral, longitudinal and combined acceleration acting on each vehicle. The two lower plots show the absolute velocity along the heading vector (slip angle is neglected at this level of detail) plotted against the time $t$ passed (middle plot) and distance traveled $s$ (lower plot). The plot against distance becomes handy when manipulating the velocity profile, since the local coordinates stay the

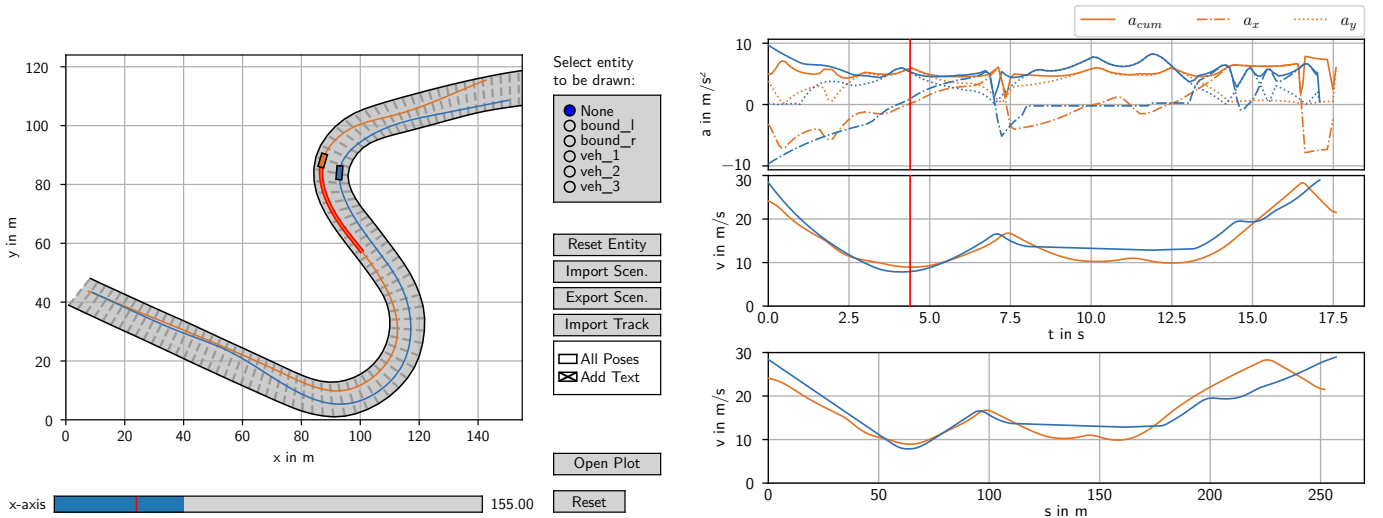

(a) Main window displaying scene at certain time-stamp. Pose of vehicles at selected time-stamp is indicated by a rectangle, the local trajectory of the VUT is highlighted in red.

(b) Temporal information window displaying course of acceleration and velocity over entire scenario horizon. Red marker selects a time-stamp to be displayed in the main window.

Fig. 1: GUI of the Scenario Architect based on Python – executable on most operating systems.

same, while the time passed is altered once the velocity is manipulated.

The scenario definition, including all technical details (track layout, vehicle paths, velocity profiles, etc.), can be carried out using a pointing device. The main window contains radio buttons which allow to select entities among single track bounds or vehicle traces. Once a radio button is selected, the corresponding path in the main window, as well as the temporal information in the second window, can be modified. Therefore, existing data points are highlighted by crosses or circles (further elaborated in Section III). Existing data points can be dragged or deleted. Clicking into the free space adds a new point to the selected entity. Since it would be cumbersome to modify every data point in the velocity plot individually, batch processing can be achieved by dragging a line while holding the left mouse button. Furthermore, dedicated buttons trigger data import and export.

When hovering with the pointing device over one of the plots in the time-window, a preview of the vehicle's motion and VUT's corresponding local trajectory at the selected time stamp intuitively supports the design of specific temporal constellations (e.g. custom design of critical situations, such as rear-ending or cut-ins). The update of the plot is fast enough to allow a synchronous scene visualization while moving the cursor on top of the temporal plot. A check box at the bottom of the main window allows the user to switch to a static view, where the vehicles are plotted with temporally equal spacing (Fig. 3). This view is especially beneficial for print media and aids the readability of complex dynamic scenarios.

### B. Track Definition

A track is defined by bounds $\boldsymbol{B}_\text{l}$ and $\boldsymbol{B}_\text{r}$, each consisting of $N_\text{B}$ coordinates $b_{\{l,r\},i}$ specifying the most left and right drivable portion of the road surface

$$\begin{aligned}\boldsymbol{B}_\text{l} &= \{\boldsymbol{b}_{\text{l},1}, \boldsymbol{b}_{\text{l},2}, ..., \boldsymbol{b}_{\text{l},N_\text{B}}\} \text{ with } \boldsymbol{b}_{\text{l},i} = \langle x_{\text{bl},i}, y_{\text{bl},i} \rangle \\ \boldsymbol{B}_\text{r} &= \{\boldsymbol{b}_{\text{r},1}, \boldsymbol{b}_{\text{r},2}, ..., \boldsymbol{b}_{\text{r},N_\text{B}}\} \text{ with } \boldsymbol{b}_{\text{r},i} = \langle x_{\text{br},i}, y_{\text{br},i} \rangle. \end{aligned} \quad (1)$$

The user may import and modify a GNSS- or LIDAR-recorded map[1] (Fig. 2) or draw a new layout from scratch. In the latter case, a sequence of coordinates can be added via the GUI for each of the bounds. The input process is supported by a live visualization displaying the generated track area and the intertwined bound points of the left and right bound. Previously drawn or imported tracks can be manipulated afterwards by dragging or removing bound points in the sequence.

---

[1]For an GNSS-recorded map, the track bound coordinates are directly determined based on satellite-localization (driving along the bounds). By contrast, a LIDAR-recorded map is an occupancy grid generated by multiple laser scans.

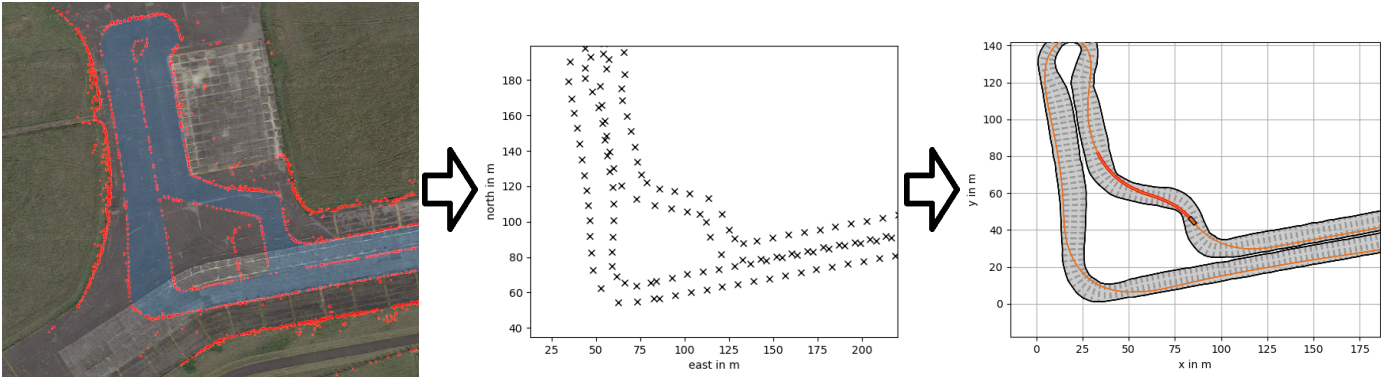

Fig. 2: Exemplary processing pipeline for an imported real-world map. An occupancy grid is generated based on LIDAR recordings (red), where bound coordinates are extracted to be imported in the Scenario Architect.

## C. Path Representation

The motion of every dynamic object in the scene is modeled by a trajectory. Each trajectory is composed of a path and an accompanying velocity profile. This information can be entered and modified separately via the GUI. The path $P_v$ of a vehicle v is based on a sequence of $N_v$ coordinates

$$P_v = \{p_{v,1}, p_{v,2}, ..., p_{v,N_v}\} \text{ with } p_{v,i} = \langle x_{v,i}, y_{v,i} \rangle. \quad (2)$$

Following previous work [14], a $C_2$ continuous cubic spline passing all points in $P_v$ is calculated for a sequence fulfilling $N_v >= 3$. Spline coefficients are determined for each neighboring pair of points in the sequence. Therefore, each point – except the first and the last – belongs to two spline segments. The $x_v$ and $y_v$ course of the $i$-th spline segment in a sequence of $N_v - 1$ spline-segments is each described by a cubic spline

$$\xi_i(\mu) = a_{3,i}\mu^3 + a_{2,i}\mu^2 + a_{1,i}\mu + a_{0,i}. \quad (3)$$

with the shared virtual path variable $\mu \in [0, 1]$. The variable $\xi$ is to be substituted by $x$ and $y$, and $f_{cs}(\cdot)$ by $\cos(\cdot)$ and $\sin(\cdot)$, respectively. The coefficients $a_i$ are chosen such that the $C_1$ and $C_2$ continuity constraint of two adjacent splines is met. The tails of the spline chain are constrained by the heading at the start $\theta_{s,1}$ and end $\theta_{e,N}$ of the path, as shown below:

$$\begin{aligned}
\xi_{s,i} &= a_{0,i} \\
\xi_{e,i} &= a_{3,i} + a_{2,i} + a_{1,i} + a_{0,i} \\
\xi'_{e,i} &= \xi'_{s,i+1} \iff 3a_{3,i} + 2a_{2,i} + a_{1,i} = \xi_{1,i+1} \\
\xi''_{e,i} &= \xi''_{s,i+1} \iff 6a_{3,i} + 2a_{2,i} = a_{2,i+1} \\
\xi'_{s,1} &= s_{\text{len},1} f_{cs}(\theta_{s,1}) = a_{1,1} \\
\xi'_{e,N} &= s_{\text{len},N} f_{cs}(\theta_{e,N}) = 3a_{3,N} + 2a_{2,N} + a_{1,N}.
\end{aligned} \quad (4)$$

The heading at the start $\theta_{s,1}$ and end $\theta_{e,N}$ of the path is determined based on the vector pointing to the neighboring point in the sequence:

$$\begin{aligned}
\theta_{s,1} &= \arctan2\left(\frac{x_{v,2} - x_{v,1}}{y_{v,2} - y_{v,1}}\right) \\
\theta_{e,N} &= \arctan2\left(\frac{x_{v,N} - x_{v,N-1}}{y_{v,N} - y_{v,N-1}}\right).
\end{aligned} \quad (5)$$

## D. Velocity Profile Initialization

The velocity initialization for every path candidate is determined based on the curvature and the corresponding friction coefficient. The maximum executable velocity is calculated based on a forward-backward solver, maxing out the combined acceleration potential, as described by Heilmeier et al. [15]. The velocity profile is updated live while a path is drawn (allowing for an accurate scenario design, e.g. enforced crashes or near-misses) and can be manipulated manually via the GUI.

## E. Export Options

Once a scenario has been laid out, it can be saved and exported. This makes it possible to load and edit previously exported scenarios. Furthermore, a structured data format for methods using the scenario data for evaluation purposes is provided (details in Section III).

Together with the Scenario Architect, we provide scripts providing a fast and easy interface for the exported scenario files. One of the scripts takes the scenario file as an input and generates either a coordinate-based map or an occupancy grid. Another script returns all dynamic parameters for the VUT and the object vehicle for a given floating-point time-stamp. Time-stamps not represented in the data-file are interpolated.

## III. Scenario-Based Testing

The entire framework, including the GUI, is implemented in Python. We demonstrate track layout generation and manipulation based on existing maps of race tracks, and based on imagination. Furthermore, the aptitude of the Scenario Architect with regard to a scenario-based validation of a safeguarding framework in an autonomous race vehicle is demonstrated.

The software architecture of the VUT follows a standard sense-plan-act principle [3]. In order to ensure the safe behavior of the trajectory planner within this structure, an online verification module is deployed [16]. We follow the principle of ASIL decomposition, as defined in the ISO 26262 [4]. In order to validate the established online verification module, we generate artificial trajectories and an object-list with the Scenario Architect. This data is then provided via the standard module interfaces. Compared to field tests, this procedure has the advantages of customized scenario design (including critical or crash scenarios) and the option of generating proper ground-truth data. Execution of such scenarios in the field would result in high costs or pose a significant safety risk.

Within this paper, we detail on the creation of three exemplary scenarios (Section III-A, III-B and III-C), each focusing on a different type of safety-critical situation. The scenarios replicate a collision of two vehicles, the ego vehicle leaving the track and an inappropriate velocity profile in a turn combination. Furthermore, we demonstrate the track creation based on real race-track data, as well as the free design by hand. Thus, these three scenarios show the wide spectrum of design possibilities. The presented scenarios are all safety-critical ones, since this is a core strategy of scenario-based testing. Nevertheless, it is also possible to design normal condition scenarios.

### A. Scenario A

The first scenario – Scenario A (Fig. 3) – is chosen to model a crash situation involving two vehicles. The idea is to have a lead vehicle (blue) that suddenly triggers an emergency brake in front of the VUT (orange). Furthermore, to increase complexity, the brake maneuver is triggered in a turn combination. Such behavior must be expected at any time by an automated vehicle. Even simple time-out problems can cause an emergency stop. Thus, this scenario can be used to challenge a safety-assessment algorithm to see if it is capable of detecting the safety-critical situation with sufficient notice. Next, we describe how this scenario is designed.

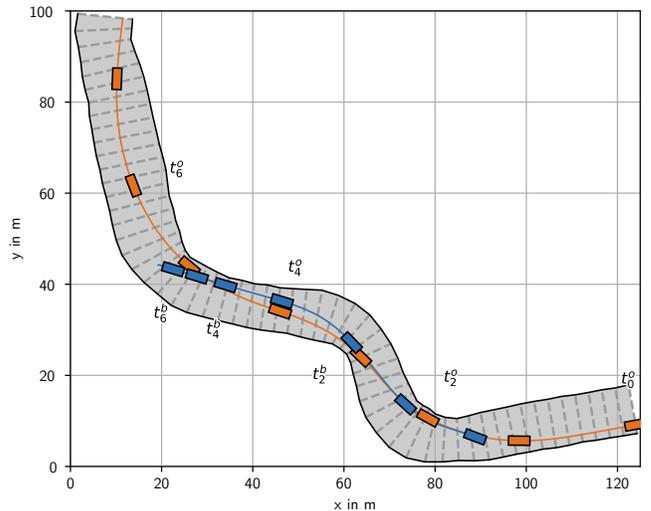

Fig. 3: Temporal overview of Scenario A, plotted using the Scenario Architect. The vehicle poses are plotted with a temporal spacing of $1s$, where some discrete time steps within this $1s$ display are indicated by $t_i^b$ and $t_i^o$.

We decided to base this scenario on a real-world track layout. The route was set up with cones on an abandoned airfield in Upper Heyford, UK – first presented in [17]. An occupation map is generated by LIDAR readings and translated into bound coordinates [18]. The bound coordinates, optionally bundled with a race-line, can then be imported by the Scenario Architect. The described work-flow, applied to this specific track, is illustrated in Fig. 2. For the scope of this scenario, only the segment holding the triple turn combination is extracted.

After importing the track limits and the race-line using the method described above, we now add an object vehicle to the scenario. First, the path of the other vehicle has to be defined, by adding support points (Fig. 4a). The first point is set at some distance from the initial pose of the VUT. The rest of the path can be chosen at will. We chose a straight segment at the end of the path, which replicates a common hard-stop maneuver. Second, the velocity profile is initialized automatically by always maximizing the available friction force. However, with the stock velocity profile, the vehicles would not result in a collision, since the other vehicle leaves the VUT's planned trajectory with sufficient notice. In order to tackle this, the live visualization of the vehicles' motion in the scene is used to synchronize and shape their intended interaction (e.g. crash at certain point in time). In order to orchestrate the motion of the vehicle, not only the path can

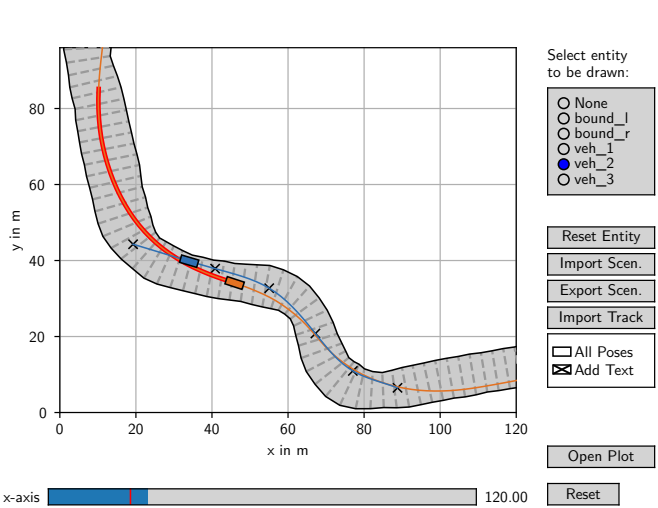

(a) Main window - implementation and modification of the object vehicle's path (blue line and black crosses).

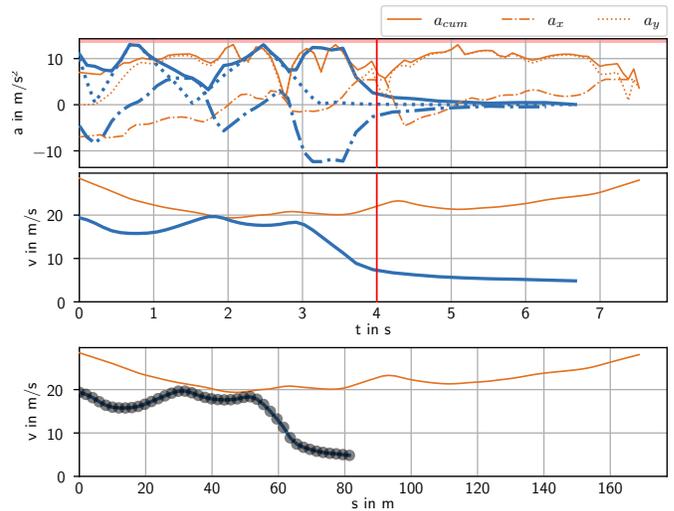

(b) Temporal information window - modification of the object vehicle's velocity profile (black dots in the lower plot).

Fig. 4: Specification of the object vehicle 'veh_2' for Scenario A. Related entities are highlighted in both windows.

be adjusted on the fly, but also the velocity profile (Fig. 4b). The customization of the velocity profile is supported by always displaying the requested pure (lateral and longitudinal) and total tire force at every time instance along the trajectory. Except for intentional fault injection, the requested tire force should always respect the friction and actuator limitations. By relying on this information, we started to decrease the velocity at $t = 3$s as fast as possible, while not exceeding the maximum combined acceleration. Once we reached $7$m/s we reduced the requested deceleration. The final trajectory pair resulted in a collision at at $t = 4.5$s.

## B. Scenario B

The second scenario – Scenario B (Fig. 5) – represents the VUT drifting towards one of the track bounds. This behavior could result from a spline-based path planner, which only checks the support points for staying within the track bounds, while the interconnecting curve bulges further outward. With this scenario, we can evaluate the capabilities of a safety system to detect a drift towards and into one of the track bounds.

By contrast to Scenario A, we did not import an existing track and instead designed one by hand. The bounds are drawn at will – here a right turn. Thereafter, the path of the VUT is designed in a way to guide it towards the bounds. The velocity profile is initialized automatically and not altered manually, since it is not key to this scenario. The step of adding further vehicles is optional and omitted here.

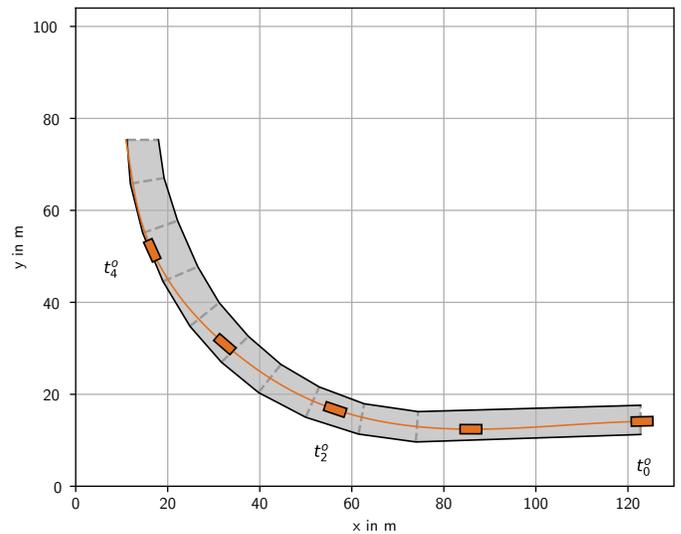

Fig. 5: Temporal overview of Scenario B, plotted with the Scenario Architect. The vehicle poses are plotted with a temporal spacing of $1s$, where some discrete time steps within this $1s$ display are indicated by $t_i^o$.

## C. Scenario C

The third scenario – Scenario C (Fig. 6) – models a VUT that plans to take a turn at too high a speed. As the centripetal force increases quadratically with speed for a fixed turn-radius, the vehicle will soon become instable. Violating the top speed in a trajectory can occur in the case of an optimization-based velocity-planner relying on slack-variables that is not able to meet the constraints for a certain turn. A safety-assessment algorithm should be able to detect such violations in good time, such that countermeasures can be applied.

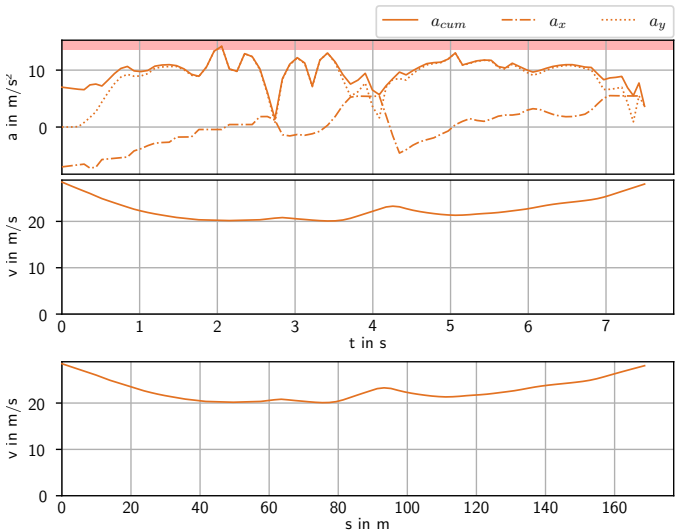

Fig. 6: Course of acceleration and velocity for Scenario C.

For this scenario, we used the exact same track section and ego-path as given in Scenario A (Fig. 3). However, the velocity profile is altered. In order to do so, we manually raised the velocity in the area around $s = 50.0$m. As a result, the velocity at the apex of the first turn is slightly higher than in Scenario A. This increase in velocity results in a higher lateral acceleration. The resulting total acceleration the tires must counteract exceeds the vehicle- and track-specific limit of $13.0$m/s$^2$.

### D. Use of Exported Scenarios

Once the scenarios are laid out and exported, they can be utilized for testing purposes. A provided script extracts data for a given floating-point timestamp. Therefore, scenarios can be played back either in real-time or fast-forwarded in order to speed up batch-testing of multiple scenarios. In that sense, the extracted data replaces the data flow of the vehicle's original functions interfacing the software under test (SUT). As a result, the SUT can be validated against the scenario's ground-truth without changes in the original software stack. In this respect, there is an analysis as to whether a developed safety framework (e.g. the framework proposed in previous work [16]) is capable of detecting all time-instances that are fault-injected in a set of scenarios. Additionally, false positives – i.e. classified failures occurring during the normal state – must be reduced by any means. Besides safety-related aspects, the Scenario Architect can be used to validate and test trajectory planners, decision managers or other software components located between perception and the control module.

Furthermore, scenarios bundled with labeled ground-truth data are well suited for CI tests. In this way, every development stage of the software is tested against a fixed set of scenarios. A successful test requires all scenarios to result in the ratings specified in the ground-truth. Otherwise, bugs and design flaws are revealed within the development process without the burden of manually testing every possible scenario constellation.

## IV. Discussion and Conclusion

The results show not only the functionality and underlying principles of the proposed Scenario Architect, but also exemplary application cases. Compared to real-world scenarios, critical scenarios including crashes can be considered with ease. To the best of the authors' knowledge, this is the first open-source tool which supports an intuitive design of critical scenarios. Within this work, we demonstrated the generation of three exemplary scenarios, each covering a distinct type of critical situation. These scenarios can be used for benchmark and approval support of safety assessment methods. Future work will focus on multi-lane support and compatibility with established scenario-description formats. The Scenario Architect will be publicly available on GitHub (github.com/TUMFTM/ScenarioArchitect).

## Acknowledgment and Contributions

Research was supported by TÜV Süd. Tim Stahl as the first author initiated the idea of this paper and is responsible for the presented concept and implementation. Johannes Betz contributed to the conception of the research project and revised the paper critically for important intellectual content. He gave final approval of the version to be published and agrees to all aspects of the work. As guarantor, he accepts responsibility for the overall integrity of the paper.